\documentclass[10pt,twocolumn,letterpaper]{article}

\usepackage{cvpr}
\usepackage{booktabs}
\usepackage{amsmath,amssymb}
\usepackage{graphicx}
\usepackage{bbm}
\usepackage{placeins}
\usepackage{float}
\definecolor{cvprblue}{rgb}{0.21,0.49,0.74}
\usepackage[pagebackref,breaklinks,colorlinks,allcolors=cvprblue]{hyperref}
\usepackage{subcaption}
\def\paperID{*****}
\def\confName{CVPR}
\def\confYear{2026}

\title{LogitDynamics: Reliable ViT Error Detection from Layerwise Logit Trajectories}

\author{Ido Beigelman\\
Technion - Israel Institute of Technology, Israel\\
{\tt\small idobeigelman@campus.technion.ac.il}
\and
Moti Freiman\\
Technion - Israel Institute of Technology, Israel\\
{\tt\small moti.freiman@bm.technion.ac.il}
}

\begin{document}
\maketitle

\begin{abstract}
Reliable confidence estimation is critical when deploying vision models. We study error prediction: determining whether an image classifier's output is correct using only signals from a single forward pass. Motivated by internal-signal hallucination detection in large language models, we investigate whether similar depth-wise signals exist in Vision Transformers (ViTs). We propose a simple method that models how class evidence evolves across layers. By attaching lightweight linear heads to intermediate layers, we extract features from the last $L$ layers that capture both the logits of the predicted class and its top-$K$ competitors, as well as statistics describing instability of top-ranked classes across depth. A linear probe trained on these features predicts the error indicator. Across datasets, our method improves or matches AUCPR over baselines and shows stronger cross-dataset generalization while requiring minimal additional computation.
\end{abstract}

\section{Introduction}

Reliable confidence estimation is crucial when vision models are deployed in high-stakes settings (e.g., clinical decision support). Despite substantial progress in uncertainty and confidence estimation, modern models can remain overconfident even when wrong, and this behavior often becomes more pronounced under distribution shift.

In this work, we study \emph{error prediction} in a Vision Transfomer (ViT) based image classifier: given an input image $x$ with label $y$, a trained classifier $f$ outputs $\hat{y}=f(x)$. Our goal is to predict the binary error indicator
\begin{equation}
e(x) = \mathbbm{1}[\hat{y} \neq y]
\end{equation}
using only \emph{internal signals} available from a single forward pass, without external verification at inference time.

Motivated by recent progress on internal-signal hallucination detection in large language models (LLMs), we ask whether analogous depth-wise signals exist in Vision Transformers. To this end, we propose a layer-dynamics method that captures how class evidence and competition evolve across depth, and uses these trajectories to predict when the model’s final prediction is likely incorrect.

\section{Background}

\paragraph{Error prediction and confidence estimation in neural networks.}
Estimating the reliability of deep neural network predictions has been widely studied in computer vision. Bayesian uncertainty estimation methods approximate the predictive posterior, including Monte Carlo (MC) dropout~\cite{gal2016dropout} and deep ensembles~\cite{lakshminarayanan2017deep}. Although effective, these approaches often incur additional computational and memory overhead at inference time. 
A complementary line of research investigates post-hoc confidence estimation using internal model signals obtained from a single forward pass. Classical approaches rely on logit and softmax-based confidence measures, including maximum logit/softmax, margin-based scores, and energy-based formulations~\cite{hendrycks2017baseline,feng2022maxlogit,liang2024selective,dadalto2024relative,liu2020energy}. Other methods operate in the feature space of deep networks, such as the Mahalanobis distance-based scoring originally proposed for out-of-distribution (OOD) detection~\cite{lee2018simple}. Beyond these techniques, several works propose training auxiliary predictors on internal representations or logits to predict misclassification~\cite{corbiere2019confidnet,aigrain2019introspectionnet}.

\paragraph{Hallucination detection in LLMs via internal signals.}
The deployment of large language models has highlighted hallucinations, where models produce confident but incorrect outputs. Recent work proposes detecting hallucinations using only internal signals (e.g., token probabilities, entropy-based measures, and internal activations), without relying on external verification \cite{azaria2023internal,orgad2025llms,barshalom2025actvit,barshalom2025signatures}. 

\paragraph{Motivation for cross-domain transfer.}
Modern vision systems and LLMs often share transformer-based backbones, suggesting that error-related internal signals (e.g., unstable belief formation across depth) may be measurable in both modalities. This motivates our evaluation of whether internal-signal hallucination detectors transfer to ViT error prediction.
\section{Methodology}

Classical confidence measures typically rely on the  logits of the classifier. While effective, such approaches ignore how class evidence evolves throughout the network. Prior work has shown that intermediate predictions can change across depth and may exhibit ``overthinking'' behavior~\cite{pmlr-v97-kaya19a}. Building on this, we propose \textbf{LogitDynamics}, a lightweight error predictor trained on depth-wise logit trajectories and top-$K$ stability features, to model \emph{layer-wise class competition dynamics} in ViTs and better predict when the model is likely to make an incorrect prediction. LogitDynamics extracts logit-based signals across layers and introduces features that capture the model's evolving class beliefs.
\subsection{Layer-wise Class Projections}
\label{sec:logit_traj__features}
Consider a pretrained ViT with $T$ transformer blocks. Let $h_t(x)$ denote the hidden representation (CLS token) in layer $t$. To expose intermediate class evidence, we attach a lightweight linear classification head to each layer. Concretely, each head maps $h_t(x)$ to class logits
\[
z^{\mathrm{head}}_t(x) \in \mathbb{R}^C,
\]
where $C$ is the number of classes. These auxiliary heads are trained to predict the ground-truth label while keeping the backbone frozen.

This procedure yields a sequence of layer-wise class-score vectors $\{z^{\mathrm{head}}_t(x)\}_{t=1}^{T}$ that approximate the model's evolving class beliefs. Let $z^{\mathrm{clf}}(x)\in\mathbb{R}^C$ denote the base model's final classifier logits and define the final predicted class as $\hat{y}=\arg\max_c z^{\mathrm{clf}}_c(x)$. From each of the last $L$ layers, we extract a $(K\!+\!1)$-dimensional vector consisting of (i) the logit $z^{\mathrm{head}}_t(x)_{\hat{y}}$ and (ii) the top-$K$ logits among classes excluding $\hat{y}$. In addition, we extract the same $(K\!+\!1)$-dimensional vector from the final classifier logits $z^{\mathrm{clf}}(x)$. We concatenate these $(L\!+\!1)$ vectors to form the primary logit-based features.

\subsection{Top-$K$ Dynamics Features}
\label{sec:top_k_dynamics_features}
Beyond raw logits, we hypothesize that \emph{instability of the model's top hypotheses across depth} is predictive of errors. We focus on the last $L$ transformer blocks and additionally include the final classifier logits as the terminal element in the sequence. Let $\{\tilde z_\ell(x)\}_{\ell=1}^{L+1}$ denote these logit vectors, where $\ell=1,\dots,L$ correspond to the last $L$ layer heads and $\ell=L+1$ corresponds to the base classifier.

For each position $\ell$, define the top-1 identity $c_\ell$ (the highest-scoring class) and the top-$K$ set $S_\ell$ (the $K$ highest-scoring classes). Using these quantities, we compute a small set of features that summarize how the model's leading hypotheses evolve across depth:

\begin{itemize}
\item \textbf{Top-1 Switch Rate:} how frequently the identity of the top-1 class changes from one depth to the next.
\item \textbf{Top-$K$ Weighted Jaccard Similarity:} how consistent the \emph{set} of top-$K$ classes remains across depth, weighting classes more when they carry more of the top-$K$ probability mass.
\item \textbf{Unique Top-$K$ Count:} how many distinct classes ever enter the top-$K$ across the depth sequence, capturing the breadth of competing hypotheses.
\item \textbf{Top-1 Mode Frequency:} how often the most common top-1 class appears across depth, measuring how strongly the model repeatedly favors a single class.
\item \textbf{Top-1 Entropy:} how dispersed the top-1 identities are across depth (low entropy indicates consistent top-1 predictions; high entropy indicates volatility).
\item \textbf{Top-1 Unique Count:} how many distinct classes appear as top-1 at any depth.
\item \textbf{Top-1 Commitment Depth:} how early the model “locks in” to its final top-1 class and maintains it for the remaining layers (earlier commitment suggests more stable decisions).
\end{itemize}

Intuitively, correct predictions tend to exhibit a stable top-ranked structure and earlier commitment, whereas errors often involve volatile competition among high-scoring classes. Formal definitions are provided in Appendix~\ref{app:dynamics_defs}.
\subsection{Error Prediction Model}

We form the final feature vector by concatenating:

\begin{enumerate}
\item the concatenation of the $(K\!+\!1)$-dimensional logit vectors extracted from each of the last $L$ layer heads together with the corresponding vector from the final classifier logits, yielding $(L\!+\!1)(K\!+\!1)$ numeric features
\item the proposed competitor-dynamics features.
\end{enumerate}
A linear classifier is trained on these features to predict the binary error indicator of the base model. Importantly, the backbone network remains frozen, and the method requires only a single forward pass at inference time.

Overall, the proposed approach preserves the efficiency of post-hoc confidence estimation while incorporating richer internal signals that capture the evolution of class competition within the network.

\section{Experimental Setup}

\subsection{Model and Datasets}

To evaluate error prediction performance, we conduct experiments on multiple image classification benchmarks. Specifically, we used the validation sets of ImageNet-1K \cite{deng2009imagenet}, CIFAR-100 \cite{krizhevsky2009cifar}, and Places365 \cite{zhou2017places}, which together provide diversity in dataset size and complexity. For all experiments, we employ a Vision Transformer (ViT-Large) \cite{dosovitskiy2021vit} as the base architecture.

\subsection{Baselines}

We compare against both classical confidence-based methods and recent internal-signal approaches inspired by hallucination detection in large language models. Let $z(x)\in\mathbb{R}^C$ denote the classifier logits over $C$ classes and $p=\mathrm{softmax}(z)$.

\subsubsection{Classical Error-Prediction Methods}

\begin{itemize}
\item \textbf{Max logit:} $s_{\text{maxlogit}}(x)=\max_{c} z_c(x)$.
\item \textbf{Entropy:} $s_{\text{ent}}(x)=-\sum_{c=1}^{C} p_c(x)\log p_c(x)$.
\item \textbf{Logit margin:} $s_{\text{margin}}(x)=z_{(1)}(x)-z_{(2)}(x)$, where $z_{(1)}\ge z_{(2)}$ are the top-2 logits.
\item \textbf{Energy score \cite{liu2020energy}:}
\[
E(x)=-T\log\sum_{c=1}^{C}\exp\!\left(z_c(x)/T\right),
\]
and we use $s_{\text{energy}}(x)=-E(x)$ so that higher scores indicate higher confidence.
\item \textbf{Mahalanobis \cite{lee2018simple}:} We follow the Mahalanobis detector of Lee et al., originally proposed for out-of-distribution and adversarial detection. The method models in-distribution training features with a class-conditional Gaussian distribution under a tied-covariance assumption. At inference, it computes layer-wise Mahalanobis scores that measure the proximity of the input feature to the nearest class mean. The resulting layer-wise scores are concatenated and a linear classifier is trained to predict errors.
\item \textbf{Top-$K$ logits:} To directly compare with our approach, we train a linear classifier on the vector of the top-$K$ logits produced by the base model's final classification head to predict the correctness of its prediction.
\end{itemize}

\subsubsection{LLM-Inspired Internal-Signal Methods}

Motivated by recent work on hallucination detection in large language models, we evaluated whether internal activation-based signals transfer to vision error prediction.

\begin{itemize}
\item \textbf{Linear probing:} We train a linear classifier on intermediate transformer representations to predict whether the model's final prediction is correct. Concretely, we extract a fixed-dimensional hidden representation (e.g., the CLS token at a chosen layer) and fit a linear model for binary error prediction.

\item \textbf{ACT-ViT \cite{barshalom2025actvit}:} We adapt ACT-ViT's core idea of \emph{learning over activation tensors}. ACT-ViT constructs an activation tensor by stacking hidden states across layers and tokens, and applies a ViT-inspired architecture over this tensor to predict hallucination/correctness, capturing global patterns across both axes.
\end{itemize}
\subsection{Metrics}

Because the error prediction task is highly class-imbalanced, we evaluate performance using the Area Under the Precision--Recall Curve (AUCPR). AUCPR provides a more informative assessment than ROC-based metrics in imbalanced settings by focusing on performance in the minority (error) class.

\section{Results}

\subsection{In-distribution performance}

We first evaluate each method under the in-distribution setting, where models are both trained and tested on the same dataset. As shown in Table~\ref{tab:indist}, LogitDynamics achieves the highest AUCPR on two of the three datasets (ImageNet and CIFAR-100) and performs comparably to the best-performing method on Places365.

Overall, logit-based classifiers consistently outperform alternative approaches. In particular, they achieve substantially better results than methods originally designed for hallucination detection in large language models, highlighting the effectiveness of logit-based uncertainty signals for misclassification detection in vision models.
\begin{table}[!htbp]
\centering
\small
\caption{In-distribution AUCPR results. $\Delta$ denotes LogitDynamics minus the best competing (second-best) method for each dataset.}\label{tab:indist}
\begin{tabular}{lccc}
\toprule
 & ImageNet & CIFAR-100 & Places365 \\
\midrule
Number of classes & 1000 & 100 & 365 \\
Dataset size & 50,000 & 10,000 & 36,500 \\
Misclassification rate & 20.34\% & 6.91\% & 44.5\% \\
\midrule
Max logit & 0.5395 & 0.3680 & 0.6969 \\
Entropy & 0.5766 & 0.3967 & 0.7038 \\
Margin & 0.5471 & 0.4197 & 0.6728 \\
Energy score & 0.4168 & 0.3580 & 0.6537 \\
Top-$K$ logits & 0.6098 & 0.4164 & \textbf{0.7283} \\
Mahalanobis & 0.3244 & 0.1064 & 0.4954 \\
ACT-ViT & 0.5390 & 0.1736 & 0.5719 \\
Linear probing & 0.5420 & 0.3050 & 0.5736 \\
LogitDynamics & \textbf{0.6458} & \textbf{0.4430} & 0.7232 \\
\midrule
LogitDynamics $\Delta$ & 0.0360 & 0.0266 & -0.0051 \\
\bottomrule
\end{tabular}
\end{table}
\subsection{Cross-dataset performance}

\begin{figure*}[!t]
  \centering
  \captionsetup{font=small}
  \captionsetup[subfigure]{font=small,skip=1pt}

  \begin{subfigure}[t]{0.46\textwidth}
    \centering
    \includegraphics[width=\linewidth]{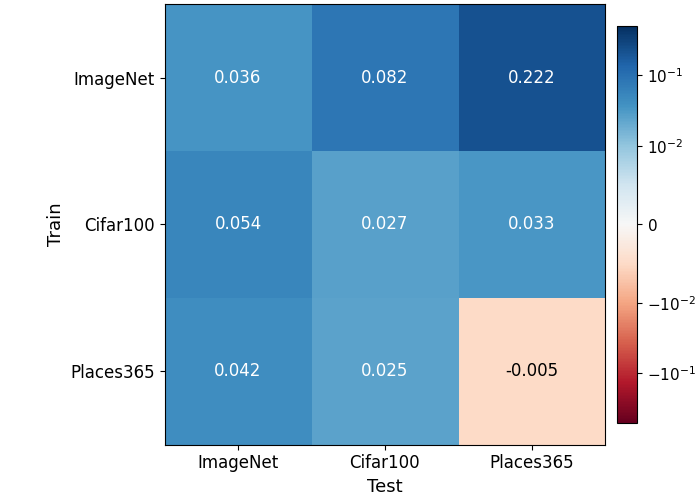}
    \caption{LogitDynamics - Top-K Logits}
  \end{subfigure}\hfill
   \begin{subfigure}[t]{0.46\textwidth}
    \centering
    \includegraphics[width=\linewidth]{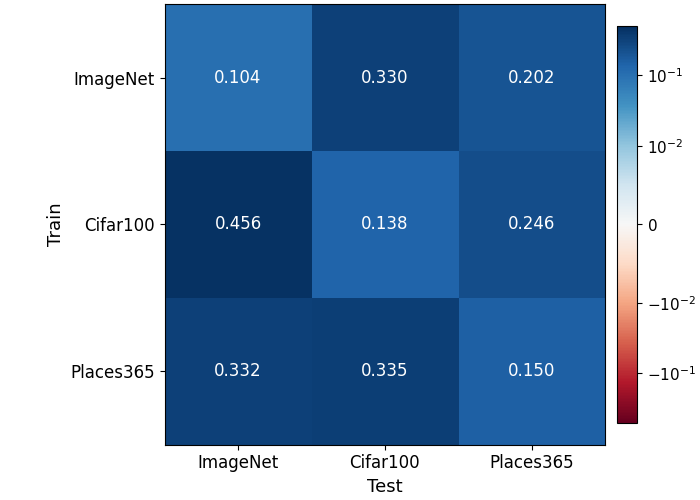}
    \caption{LogitDynamics - Linear probing}
  \end{subfigure}\hfill

  \vspace{-0.3em}
\caption{Cross-dataset AUCPR for each baseline method, reported as the performance difference relative to LogitDynamics (LogitDynamics $-$ method). Rows indicate the training dataset; columns indicate the test dataset.}
  \label{fig:cross_dataset_all}
\end{figure*}

We next evaluate cross-dataset generalization using AUCPR differences measured relative to LogitDynamics. For each method, we train the error predictor on one dataset and evaluate it on all three datasets, producing a complete train--test matrix of AUCPR differences relative to LogitDynamics. In Fig.~\ref{fig:cross_dataset_all}, the diagonal entries reflect in-domain gaps, while off-diagonal entries measure how each method’s performance changes under distribution shift compared to LogitDynamics, providing a robustness view beyond in-distribution evaluation.

Overall, linear probing exhibits a pronounced decline in cross-dataset performance relative to LogitDynamics. ACT-ViT demonstrates a similar trend (see Appendix~\ref{app:actvit_cross} and Fig.~\ref{fig:cross_dataset_actvit}). In contrast, logit-based signals show comparatively smaller performance degradations, indicating stronger transferability in vision settings. Among all methods, LogitDynamics consistently exhibits the smallest performance drop under cross-dataset transfer.

\subsection{Ablations}
\begin{figure}[t]
  \centering
  \includegraphics[width=\linewidth]{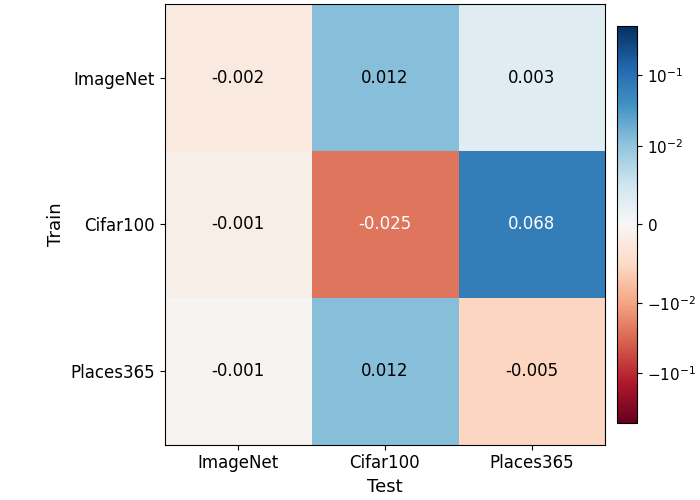}
\caption{\textbf{Ablation study.} Cross-dataset AUCPR differences reported as \emph{with} Top-$K$ dynamics features minus \emph{without} Top-$K$ dynamics features. Positive values indicate that adding Top-$K$ dynamics improves AUCPR, while negative values indicate a decrease.}
  \label{fig:ablation}
\end{figure}

We ablate the contribution of the Top-$K$ dynamics statistics (Sec.~\ref{sec:top_k_dynamics_features}) beyond the logit trajectory features (Sec.~~\ref{sec:logit_traj__features}). Fig.~\ref{fig:ablation} reports
$\text{AUCPR}(\text{w dynamics})-\text{AUCPR}(\text{w/o dynamics})$,
so positive values indicate that adding dynamics features improves AUCPR.

Overall, dynamics features provide little benefit in-distribution but improve cross-dataset transfer: the mean diagonal difference (train$=$test) is $-0.0107$, while the mean off-diagonal difference (train$\neq$test) is $0.0155$. We hypothesize these features capture stability of class competition under distribution shift, acting as a robustness signal, while adding mild variance in-domain where layer logits are already highly informative.

\section{Conclusion}

We studied error prediction in Vision Transformers using internal signals available from a single forward pass. Motivated by hallucination-detection methods in large language models, we investigated whether internal signals can improve reliability estimation in vision models. We proposed a simple approach that models the dynamics of class evidence across layers by combining layer-wise logit features with statistics capturing instability among the top-ranked classes. 

Experiments across ImageNet-1K, CIFAR-100, and Places365 show that LogitDynamics consistently improves or matches AUCPR over classical logit-based confidence measures and internal-activation baselines, while requiring minimal additional computation and no modification to the backbone network. These results suggest that depth-wise belief dynamics provide a useful and complementary signal for predicting model errors. Future work may explore extending these signals to other architectures, tasks, and distribution shifts.
\clearpage

{\small
\bibliographystyle{ieeenat_fullname}
\bibliography{main}
}

\clearpage
\appendix
\section{Appendix}
\subsection{Additional Cross-Dataset Results}
\label{app:actvit_cross}

ACT-ViT mirrors the behavior of linear probing, demonstrating poor cross-dataset generalization and underperforming LogitDynamics across train--test combinations.

\begin{figure}[H]
  \centering
  \includegraphics[width=\linewidth]{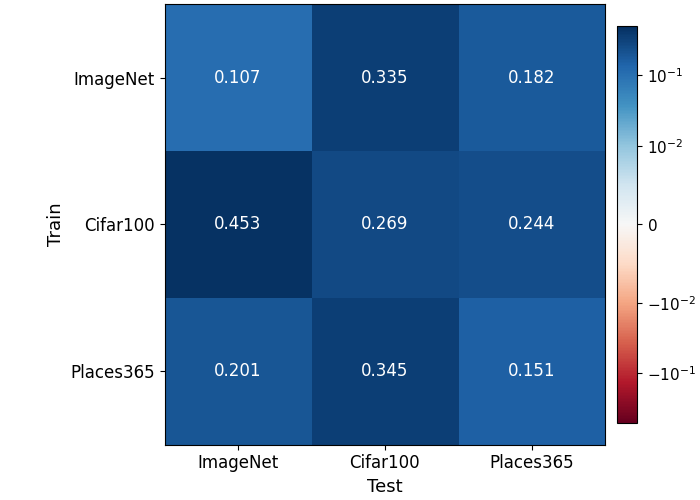}
  \caption{Cross-dataset AUCPR differences relative to LogitDynamics for ACT-ViT. Rows: training dataset; columns: test dataset.}
  \label{fig:cross_dataset_actvit}
\end{figure}

\subsection{Top-$K$ Dynamics Feature Definitions}
\label{app:dynamics_defs}

This appendix provides the formal definitions for the Top-$K$ dynamics features described in Sec.~3.2.

\paragraph{Depth-wise logit sequence.}
We focus on the last $L$ transformer blocks and additionally include the base classifier logits as the terminal element. Let $T$ be the total number of transformer blocks. Define the sequence $\{\tilde z_\ell(x)\}_{\ell=1}^{L+1}$ by
\[
\tilde z_\ell(x)=
\begin{cases}
z^{\mathrm{head}}_{T-L+\ell}(x), & \ell=1,\dots,L,\\
z^{\mathrm{clf}}(x), & \ell=L+1,
\end{cases}
\]
where $z^{\mathrm{head}}_{t}(x)\in\mathbb{R}^C$ denotes the logits produced by the auxiliary linear head attached to layer $t$, and $z^{\mathrm{clf}}(x)\in\mathbb{R}^C$ denotes the base model’s final classifier logits.

\paragraph{Top-1 and Top-$K$ identities.}
For each $\ell$, define the top-1 identity
\[
c_\ell=\arg\max_{c\in\{1,\dots,C\}} \tilde z_{\ell,c}(x),
\]
and the top-$K$ set
\[
S_\ell=\mathrm{TopK}\big(\tilde z_\ell(x)\big)\subseteq\{1,\dots,C\},
\]
where $\mathrm{TopK}$ returns the indices of the $K$ largest components.

\paragraph{Restricted-softmax weights over Top-$K$.}
We define normalized weights over the top-$K$ logits by applying a softmax restricted to $S_\ell$:
\[
w_\ell(i)=
\begin{cases}
\displaystyle \frac{\exp(\tilde z_{\ell,i}(x))}{\sum_{j\in S_\ell}\exp(\tilde z_{\ell,j}(x))}, & i\in S_\ell,\\[6pt]
0, & i\notin S_\ell.
\end{cases}
\]
In practice, we compute this stably by subtracting $\max_{j\in S_\ell}\tilde z_{\ell,j}(x)$ inside the exponent.

\paragraph{Top-1 Switch Rate.}
\[
\mathrm{SR}(x)=\frac{1}{L}\sum_{\ell=1}^{L}\mathbbm{1}\!\left[c_\ell\neq c_{\ell+1}\right].
\]

\paragraph{Top-$K$ Weighted Jaccard Similarity.}
For adjacent depths $(\ell,\ell+1)$ define the weighted intersection mass
\[
I_{\ell,\ell+1} \;=\; \sum_{i\in S_\ell\cap S_{\ell+1}} \min\{w_\ell(i),w_{\ell+1}(i)\}.
\]
The weighted Jaccard similarity is
\[
J_w(S_\ell,S_{\ell+1})
\;=\;
\frac{I_{\ell,\ell+1}}{\sum_{i\in S_\ell} w_\ell(i) + \sum_{i\in S_{\ell+1}} w_{\ell+1}(i) - I_{\ell,\ell+1}}.
\]
We report the mean over adjacent pairs:
\[
\frac{1}{L}\sum_{\ell=1}^{L} J_w(S_\ell,S_{\ell+1}).
\]

\paragraph{Unique Top-$K$ Count.}
\[
\left|\bigcup_{\ell=1}^{L+1} S_\ell\right|.
\]

\paragraph{Top-1 Mode Frequency.}
Let $n(c)=\sum_{\ell=1}^{L+1}\mathbbm{1}[c_\ell=c]$ be the count of each top-1 identity and define $p(c)=n(c)/(L+1)$. The mode frequency is
\[
\max_{c} p(c).
\]

\paragraph{Top-1 Entropy.}
Using the same empirical distribution $p(c)$ over top-1 identities,
\[
-\sum_{c} p(c)\log p(c).
\]

\paragraph{Top-1 Unique Count.}
\[
\left|\{c_\ell\}_{\ell=1}^{L+1}\right|.
\]

\paragraph{Top-1 Commitment Depth.}
Let $c_{L+1}$ be the final element's top-1 identity. Define the earliest depth at which the top-1 remains equal to the final top-1 for all subsequent depths:
\[
\ell^\star=\min\{\,\ell\in\{1,\dots,L+1\}: c_\ell=c_{\ell+1}=\cdots=c_{L+1}\,\}.
\]
We normalize the commitment depth as
\[
\frac{\ell^\star-1}{L},
\]
which lies in $[0,1]$ with smaller values indicating earlier commitment.

\subsection{Data Splits and Hyperparameters}
\label{app:splits_hparams}

\paragraph{Splits.}
For each dataset with $N$ examples, we construct stratified splits with respect to the binary error label $e(x)$ (incorrect vs.\ correct). We first split the data into an $85\%$ train/validation partition and a $15\%$ held-out test set. The $85\%$ partition is further split into: (i) a \emph{head-training} subset used to train the per-layer linear classification heads, and (ii) a \emph{probe pool} used to train and validate the error predictor. Concretely, we allocate a fraction $p_{\text{probe}}$ of the $85\%$ partition to the probe pool (default $p_{\text{probe}}=0.2$), and the remaining portion to head training. For CIFAR-100, we use a larger probe fraction ($p_{\text{probe}}=0.3$) due to its smaller sample size and stronger class imbalance in the error labels. Finally, we split the probe pool into $75\%$ probe-train and $25\%$ probe-val. All stages use stratified sampling by $e(x)$.

\paragraph{Auxiliary layer heads.}
We attach a linear head to each transformer block in a suffix of the network (the last $L$ blocks). Heads are trained with cross-entropy on the head-training split while keeping the ViT backbone frozen. We optimize with AdamW, using a batch size of $512$ and weight decay $0.0$. In hyperparameter sweeps, we vary the head learning rate in $\{10^{-4}, 2\cdot 10^{-4}, 5\cdot 10^{-4}, 7\cdot 10^{-4}, 10^{-3}\}$ and the number of head-training epochs in $\{2,5,7,10,12,16\}$.We select the best configuration based on probe-val AUCPR. For cross-dataset generalization experiments, we extract features using auxiliary heads trained on the \emph{target} dataset (i.e., we train the auxiliary heads on the target dataset's training split with a frozen backbone, and then evaluate the transferred error predictor on target features).

\paragraph{Feature construction.}
Unless stated otherwise, we use the last $L\in\{1,3,5,7,9,12,16,20,24\}$ transformer blocks and include the base classifier logits as the terminal element in the depth sequence. We extract numeric logit features using the predicted-class logit together with the top-$K$ competitor logits, with $K\in\{1,3,5,7,10\}$; competitors exclude the predicted class for the numeric logit features. Dynamics features (Sec.~3.2) are computed using the raw top-$K$ sets at each depth (without excluding the predicted class). We  include the base classifier logits as an additional block of numeric features.

\paragraph{Normalization.}
All features are standardized using the mean and standard deviation computed on the \emph{probe-train} split only, and we apply the same normalization to probe-val and the held-out test set to avoid leakage. For cross-dataset generalization experiments, we compute normalization statistics using the \emph{target dataset's training split}, and apply this normalization when evaluating on the target dataset.

\paragraph{Error predictor.}
The error predictor is a linear classifier trained with AdamW on the probe-train split. We train for $100$ epochs with learning rate $10^{-3}$, batch size $256$. Because the error label is imbalanced, we use a positive-class weighting term in the binary cross-entropy loss, where the positive weight is set to $\frac{N_{\text{neg}}}{N_{\text{pos}}}$ computed on the probe-train split.

\paragraph{Model selection.}
We select hyperparameters using probe-val AUCPR (PR-AUC). Final results are reported on the held-out $15\%$ test set using the same trained heads, normalization statistics, and error predictor.

\end{document}